\documentclass{article}

\usepackage[numbers]{natbib}
\usepackage[final]{nips_2016}
\usepackage[utf8]{inputenc} % allow utf-8 input
\usepackage[T1]{fontenc}    % use 8-bit T1 fonts
\usepackage{hyperref}       % hyperlinks
\usepackage{url}            % simple URL typesetting
\usepackage{booktabs}       % professional-quality tables
\usepackage{amsfonts}       % blackboard math symbols
\usepackage{nicefrac}       % compact symbols for 1/2, etc.
\usepackage{microtype}      % microtypography

\usepackage{graphicx}
\usepackage{algorithm}
\usepackage[noend]{algpseudocode}
\algnewcommand\algorithmicinput{\textbf{input:}}
\algnewcommand\INPUT{\item[\algorithmicinput]}
\algnewcommand\algorithmicoutput{\textbf{output:}}
\algnewcommand\OUTPUT{\item[\algorithmicoutput]}
\usepackage{amsmath,amssymb}
\DeclareMathOperator{\E}{\mathbb{E}}

\title{Combination of Hyperband and Bayesian Optimization for Hyperparameter Optimization \\ in Deep Learning}

\author{
  Jiazhuo Wang \\ 
  Blippar \\
  \texttt{jiazhuo.wang@blippar.com}
  \And Jason Xu \\
  Blippar \\
  \texttt{jason.xu@blippar.com}
 \And Xuejun Wang \\
  Blippar \\
  \texttt{xuejun.wang@blippar.com}
}

\begin{document}

\maketitle

\begin{abstract}

Deep learning has achieved impressive results on many problems. However, it requires high degree of expertise or a lot of experience to tune well the hyperparameters, and such manual tuning process is likely to be biased. Moreover, it is not practical to try out as many different hyperparameter configurations in deep learning as in other machine learning scenarios, because evaluating each single hyperparameter configuration in deep learning would mean training a deep neural network, which usually takes quite long time. Hyperband algorithm achieves state-of-the-art performance on various hyperparameter optimization problems in the field of deep learning. However, Hyperband algorithm does not utilize history information of previous explored hyperparameter configurations, thus the solution found is suboptimal. We propose to combine Hyperband algorithm with Bayesian optimization (which does not ignore history when sampling next trial configuration). Experimental results show that our combination approach is superior to other hyperparameter optimization approaches including Hyperband algorithm.

\end{abstract}

\section{Introduction}
We have been witnessing the breakthrough brought by deep learning in various research topics: Computer vision, natural language processing, and speech recognition; and the significant impact on many industries: Augmented reality, self driving cars, and smart home devices, to name a few. However, the design of deep neural networks usually requires huge amount of efforts even for experienced deep learning researchers and practitioners, due to the high complexity involved in nature with these deep neural networks (for example, some modern deep neural networks may even have thousands of layers).  What makes the adoption and application of a deep learning based solution even more complex is that (1) all feasible hyperparameter configurations form a huge space, from which we need to choose the optimal case, and that (2) evaluating even a single hyperparameter configuration point in the space is often very time consuming (i.e., training one deep neural network using modern GPU usually takes hours, days, or even weeks). 

Hyperparameters in deep learning can be roughly divided into two types. One type is those associated with the learning algorithms. For example, suppose one has chosen an off-the-shelf deep neural network, such as AlexNet, or ResNet, or one has designed a new network specifically for his or her own problem; when it comes to training the network, it is usually not clear in advance, for the training dataset in hand, what learning rate is appropriate, after how many epochs we should decrease the learning by how much, what value we need to set for weight decay, and so on. Setting these hyperparameters correctly is often critical for reaching the full potential of the deep neural network chosen or designed.

The other type of hyperparameters in deep learning is related to how we design the deep neural networks. For example, some important design questions include: How many layers we need for our network, how many filters a given convolutional layer needs, at which layer we downsample the size of feature maps, and so on. Choosing different values for these hyperparameters would result into very different network architectures, and consequently would influence the quality of the produced results by them.

In the deep learning community, choosing appropriate values for the above two types of hyperparameters is usually accomplished manually, which largely depends on one's own experience or intuition, and it may be hard to reproduce results of similar quality when others would like to apply on different datasets or problems. Moreover, such manual tuning makes it difficult to understand the real ability of the deep learning model: One may overestimate or underestimate the model if one just happens to choose a good or bad hyperparameter configuration by chance.

A more systematic approach to tune hyperparameters usually adopted in machine learning community is grid search. However, when applied to certain hyperparameter optimization problems in deep learning, it is too time-consuming. For example, suppose one has decided to design a network with 20 convolutional layers, and the hyperparameters are the number of filters to be assigned to each of the 20 convolutional layers (i.e., we have 20 hyperparameters). If the number of filters is restricted in certain range, e.g., between 10 and 512, then there would be $(512-10)^{20} = 502^{20}$ points in total in the hyperparameter configuration space; even if grid length is set as 10, there are still $(\frac{502}{10})^{20} = 50^{20}$ points to be evaluated. This would take unacceptable long time: Remember evaluating each single point means training a network till convergence.

Random search \cite{Bergstra:2012:RSH:2188385.2188395} proposes to randomly sample points from the hyperparameter configuration space. Although this approach looks simple, but it is shown to be able to find comparable hyperparameter configuration to grid search in deep learning, but using only a small fraction of the computation time used by grid search. Granted with the same computational budget, random search can find better configurations.

The state-of-the-art Hyperband algorithm \cite{DBLP:journals/corr/LiJDRT16}  makes hyperparameter optimization much more efficient by observing that it is usually not necessary to allocate computational budget or resources uniformly to every hyperparameter configuration; a smarter way is to allocate more budget to more promising hyperparameter configurations while eliminating the less promising ones quickly. When applied to deep learning, this algorithm can be viewed as an early-stopping strategy that one often adapts when tuning hyperparameters for neural networks training: If a learning curve does not look quite promising or even very poor, one would be more likely to abandon the current hyperparameter configuration. Further more, it does early-stopping in a more principled way, as would be briefly reviewed in the Methodology Section. 

One major limitation of Hyperband algorithm is that it treats all hyperparameter configuration points as being independent from each other. But the hyperparameter configuration space is usually smooth, which means that knowing the quality of certain points in it might help infer the quality of their nearby points. Therefore, it would be more efficient if we do not sample trial points independently, but according to what we have known before (i.e., the quality of all previously sampled trial points). Actually, Bayesian optimization \cite{Shahriari:2015, NIPS2011_4443,DBLP:journals/corr/abs-1209-5111} utilizes the above smoothness assumption. When sampling the next trial point, it balances between exploitation and exploration. Exploitation means it samples from regions where good quality trial points were found nearby before, and exploration means it also examines regions that have been rarely visited before. But we notice that unlike Hyperband algorithm, Bayesian optimization still allocates computational budget uniformly, which causes efficiency issues.

 To find better hyperparameter configurations as efficiently as possible, we propose to combine Hyperband algorithm and Bayesian optimization. We are motivated by the fact that the two methods are complementary in nature with each other. After combining them, we are able to fully utilize what we learned through history of all previous trial points, and at the same time, to focus our attention on really promising ones and avoid wasting time on less promising ones. In a nutshell, the spirit of our combination approach is quite straightforward: We simply follow the entire process of Hyperband algorithm, but one critical difference is that instead of sampling the set of trial points independently, we sample each of them sequentially (i.e., sampling order matters) according to Bayesian optimization criterion: Balancing between exploitation and exploration. 

Experimental results show that our combination approach outperforms state-of-the-art Hyperband algorithm, on several deep learning problems and hyperparameter optimization scenarios. The results also indicate that the more difficult the hyperparameter optimization problem is, the larger the margin is between our approach and other approaches.                          

\section{Methodology}
In this section, we present our core algorithm. We first briefly review the critical ideas that are helpful for understanding our combination approach, respectively in Hyperband algorithm and Bayesian optimization. Then we illustrate in details why and how we combine them together.

\subsection{Hyperband}
The essential idea of Hyperband is to allocate more resources to more promising hyperparameter configurations. First, it initializes a set of ${n}$ trial points (each trial point corresponding to one hyperparameter configuration). Then, it uniformly allocates a budget to each trial point, and evaluates its performance (i.e., objective function) given that budget. A proportion of trial points with worse performance are deemed as to be less promising, and thus are removed from the set (namely, now there are less trial points in the set). This procedure repeats several rounds until only a single trial point remains in the set. 

This idea is based on the assumption that a trial point's performance obtained at earlier rounds is a good indicator of its final performance. Apparently, the more resources each trial point is allocated, the more reliable this indicator is. Given a total finite budget $B$ of resources, roughly each trial point is allocated $B / n$ resources. If a large $n$ is used, each trial point would be allocated less resources, thus earlier rounds performance may not reflect well the final performance. On the contrary, if a small $n$ is used, although each trial point is more likely to be allocated sufficient resources to reflect its final performance, the algorithm would suffer from a pool of less candidates. To alleviate this dilemma, Hyperband actually tries out different values for $n$, and returns the best hyperparameter configuration among them. 

Hyperband algorithm is presented in Algorithm~\ref{alg:hyperband} (see \cite{DBLP:journals/corr/LiJDRT16} for details). The function $get\_hyperparameter\_configuration(n)$ returns a set of $n$ i.i.d. trial points from the predefined configuration space. The function $run\_then\_return\_obj\_val(x, r)$ first trains the trial point $x$ with $r$ amount of resources, and then returns the objective function value $f(x)$ evaluated after training. And the function $top\_k(trials, obj\_vals, k)$ first ranks all the input $trials$ according to their objective function value $obj\_vals$, and then returns the top $k$ performing trial points.

\begin{algorithm}
\caption{Hyperband algorithm} \label{alg:hyperband}
\begin{algorithmic}[1]
\INPUT maximum amount of resource that can be allocated to a single hyperparameter configuration $R$, and proportion controller $\eta$
\OUTPUT one hyperparameter configuration
\State \textbf{initialization:} $s_{max} = \left \lfloor log_{\eta}(R) \right \rfloor, \; B = (s_{max} + 1)R$
\For {$s \in \{s_{max}, s_{max}-1, ..., 0 \} $}
 	\State $n = \left \lceil \frac{B}{R} \frac{\eta^{s}}{(s+1)} \right \rceil$, $r = R\eta^{-s}$
	\State $X = $ get\_hyperparameter\_configuration($n$)
	\For {$i \in {0, ..., s}$}
		\State $n_{i} = \left \lfloor n\eta^{-i} \right \rfloor$
		\State $r_{i} = r\eta^{i}$
		\State $F = $ \{ run\_then\_return\_obj\_val($x, r_{i}$) : $x \in X$\}
		\State $X = $ top\_k($X, F, \left \lfloor n_{i} / \eta \right \rfloor$)
	\EndFor
\EndFor
\Return configuration with the best objective function value
\end{algorithmic}
\end{algorithm}

\subsection{Bayesian Optimization}
The main idea of Bayesian optimization is as follows. Since evaluating the objective function $f$ for a trial point $x$ is very expensive, it approximates $f$ using a probabilistic surrogate model that is much cheaper to evaluate, and it is an iterative process: (1) The algorithm samples the next trial point $x_{t+1}$ according to the current surrogate model. (2) It evaluates $f(x_{t+1})$. (3) It updates the surrogate model based on the new data point $(x_{t+1}, f(x_{t+1}))$, and goes back to step (1). Bayesian optimization samples trial points sequentially, and each trial point is sampled utilizing all the information in the history (reflected by the built surrogate model). Namely, what will be sampled next is actually determined by all of those sampled previously.

Bayesian optimization is presented in Algorithm~\ref{alg:bayesion-optimization}. On line~\ref{alg:line:acquisition-func}, the next trial point is sampled at the place optimizing an acquisition function $\mu(x|D_{t})$. The expected improvement in Equation \ref{eq:acquisition-func} is usually used as the acquisition function in the literature, where $x^{+}$ is the trial point with the best objective function value found in the first $t$ steps. Note that the expected improvement acquisition function favors points with high variance and high mean value. Namely, it balances between exploration of the not well explored regions and exploitation of the visited regions, in the configuration space. 

\begin{equation} \label{eq:acquisition-func}
\mu(x|D_{t}) = \E(max\{0, f_{t+1}(x) - f(x^{+})\} | D_{t})
\end{equation}

In particular, two types of probabilistic models have been proposed as the surrogate to compute the expectation in Equation \ref{eq:acquisition-func}. Discriminative models like Gaussian process (GP) directly models $p(f(x)|x)$, while generative models like Tree-structured Parzen Estimator (TPE) models $p(x|f(x))$ and $p(f(x))$. As reported \cite{NIPS2011_4443}, TPE achieves better results than GP.

\begin{algorithm}
\caption{Bayesian optimization} \label{alg:bayesion-optimization}
\begin{algorithmic}[1]
\State \textbf{initialization:} $D_{0} = \emptyset$
\For {$t \in \{ 1, 2, ... \} $}
	\State $x_{t+1} = argmax_{x}\mu(x|D_{t})$ \label{alg:line:acquisition-func}
	\State Evaluate $f(x_{t+1})$
	\State $D_{t+1} = D_{t} \cup \{(x_{t+1}, f(x_{t+1}))\}$
	\State Update probabilistic surrogate model using $D_{t+1}$
\EndFor
\end{algorithmic}
\end{algorithm}

\subsection{Combination of Hyperband and Bayesian Optimization}
Although both Hyperband and Bayesian Optimization are very powerful, each has its own strength and weakness. We observe that the two approaches are actually complementary to each other, namely, the weakness of one is the strength of the other, and vice versa, as illustrated next. Therefore, we propose to combine them together to obtain an enhanced hyperparameter optimization algorithm.

\begin{algorithm}
\caption{Combination of Hyperband and Bayesian optimization} \label{alg:hyperband-bayesion-optimization}
\begin{algorithmic}[1]
\INPUT maximum amount of resource that can be allocated to a single hyperparameter configuration $R$, and proportion controller $\eta$
\OUTPUT one hyperparameter configuration
\State \textbf{initialization:} $s_{max} = \left \lfloor log_{\eta}(R) \right \rfloor, \; B = (s_{max} + 1)R$
\For {$s \in \{s_{max}, s_{max}-1, ..., 0 \} $}
 	\State $n = \left \lceil \frac{B}{R} \frac{\eta^{s}}{(s+1)} \right \rceil$, $r = R\eta^{-s}$
	\For {$i \in {0, ..., s}$}
		\State $n_{i} = \left \lfloor n\eta^{-i} \right \rfloor$
		\State $r_{i} = r\eta^{i}$
		\If {$i == 0$} \label{alg:line:begin-first-round}
			\State $X = \emptyset$, $D_{0} = \emptyset$
				\For {$t \in \{ 1, 2, ..., n_{i} \} $}
					\State $x_{t+1} = argmax_{x}\mu(x|D_{t})$
					\State $f(x_{t+1}) = $  run\_then\_return\_obj\_val($x, r_{i}$)
					\State $X = X \cup \{x_{t+1}\}$
					\State $D_{t+1} = D_{t} \cup \{(x_{t+1}, f(x_{t+1}))\}$
					\State Update probabilistic surrogate model using $D_{t+1}$ \label{alg:line:end-first-round}
				\EndFor
		\Else
			\State $F = $ \{ run\_then\_return\_obj\_val($x, r_{i}$) : $x \in X$\}
			\State $X = $ top\_k($X, F, \left \lfloor n_{i} / \eta \right \rfloor$)
		\EndIf
	\EndFor
\EndFor
\Return configuration with the best objective function value
\end{algorithmic}
\end{algorithm}

More specifically, on one hand, all the trial points in Hyperband are sampled independently. However, once a trial point $x$ is sampled and the objective function $f(x)$ at that trial point is evaluated, this data point $(x, f(x))$ tells us something about how our objective function behaves on the configuration space. Intuitively, if $f(x)$ is a bad one, we should avoid sample trial points near $x$ subsequently; otherwise, we should consider sampling more trial points near $x$. Note that it is only reasonable to take these actions under the condition that the objective function is smooth over the configuration space, which is usually a weak assumption (i.e., easy to be satisfied) for a wide spectrum of hyperparameter optimization problems encountered in deep learning. Each trial point in Hyperband is sampled independently from all the trial points sampled previously, thus it fails to utilize the lessons learned in the history, which may result into wasting precious time on repeating errors that already happened in history.

On the other hand, in Bayesian optimization, each trial point $x$ is allocated sufficient resources to obtain the final objective function value $f(x)$. In deep learning, allocating sufficient resources to a single trial point usually means training a deep neural network to convergence, which may take hours, days, or even weeks using modern GPU. In this case, one cannot afford allocating whatever amount of resources required by a single trial point to obtain its final objective function value, without knowing in advance that this value is likely to be the best one among those of all the trial points.

From the above analysis of the weakness of each approach, we see that they are actually perfectly complementary to each other. Trial points in Hyperband should be sampled sequentially, and each trial point should be sampled according to the experience obtained at all previously sampled and evaluated trial points, which is exactly how Bayesian optimization works. An early stopping like mechanism as used in Hyperband should be adopted in Bayesian optimization, so that trials points that are already worse than other trial points at early rounds should be abandoned, since their final performance is also less likely to be the best among all.

We propose to combine Hyperband and Bayesian optimization in a straightforward manner as follows. We generally follow the process of Hyperband algorithm. However, the only difference is that, instead of sampling all $n$ trial points independently at once as in Hyperband, in our combination approach, we sample trial points one by one using Bayesian optimization. More specifically, in each first round of hyperband (on different initial value of $n$), once a trial point $x_{t+1}$ is sampled, its intermediate performance $f(x_{t+1})$ given the current allocated resources $r_{0}$ is evaluated. This new data point $(x_{t+1}, f(x_{t+1}))$ contributes to updating the surrogate probabilistic model in Bayesian optimization, based on which the next trial point will be sampled.

As illustrated in Algorithm~\ref{alg:hyperband-bayesion-optimization}, the bulk of our combination algorithm is very similar to Hyperband in Algorithm~\ref{alg:hyperband}, and the major difference is from line~\ref{alg:line:begin-first-round} to line~\ref{alg:line:end-first-round}, which are basically Bayesian optimization in Algorithm~\ref{alg:bayesion-optimization}.

\section{Experiments}
In this section, we empirically evaluate the performance of our combination algorithm. We conduct two major sets of experiments. The experiments in the first set are those related to deep learning problems mentioned in the Hyperband paper. In the second set, there is only one experiment (but much more complicated and harder than those in the first set), whose target is to find a decomposed neural network architecture of SSD \cite{DBLP:journals/corr/LiuAESR15} (one of the state-of-the-art object detectors) balancing well the accuracy and speed for object detection. All our experiments are conducted in the Caffe framework.

In our combination algorithm, we choose TPE \cite{NIPS2011_4443} as the Bayesian optimization part, due to its supreme performance than its discriminative counterpart GP. In all the experiments, we compare the performance of our combination algorithm, named as Hyperband\_TPE, against three baselines: Random search, TPE, and Hyperband.

\subsection{LeNet and AlexNet}
In this section, we repeat the experiments (four in total) related to deep learning problems in the Hyperband paper. We strictly follow the setting up protocols in the paper, unless otherwise specified below (which is largely due to that we want to restrict the total time taken by any approach to be no more than a week on our two Titan X GPUs).

\subsubsection{LeNet on MNIST}
The experiment in this section is to train LeNet on MNIST dataset using mini-batch SGD. There are four hyperparameters: Learning rate, batch size, and number of filters for the two convolutional layers of LeNet. The range of feasible values each hyperparameter could take follows that in the Hyperband paper. 

\begin{figure}
  \centering
    \includegraphics[width=0.6\textwidth]{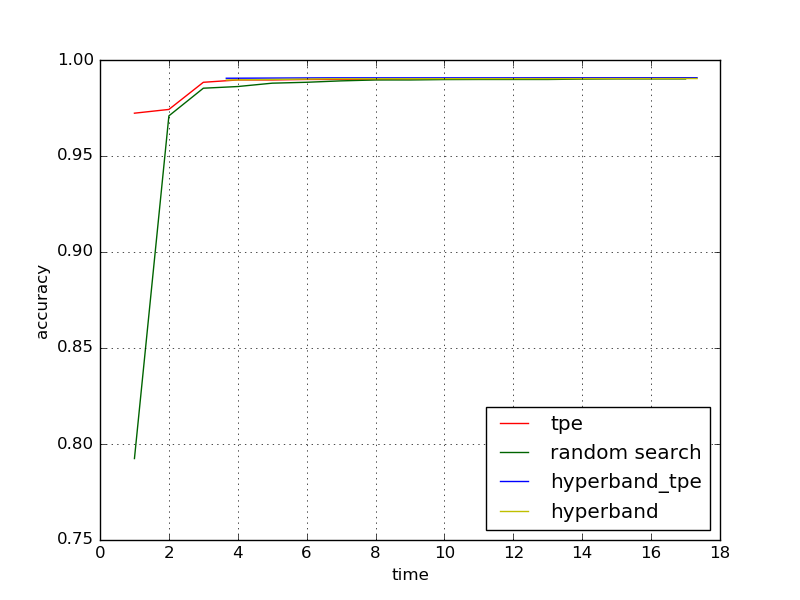}
  \caption{Average accuracy across 10 trials for LeNet on MNIST.}
  \label{fig:mnist}
\end{figure}

For Hyperband and our combination algorithm, we set $R$ = 81, and $\eta$ = 3, the same as in the Hyperband paper; and we empirically set up the number of training images corresponding to each unit of $R$ (i.e., one of the 81 pieces in this particular setting), which is not mentioned in the Hyperband paper. These settings result into an equivalent number (in the sense of total time taken) of about 18 trial points for either TPE or random search, where each trial point consumes all 81 units of $R$. In Hyperband paper, to make sure the statistical significance of the results, it runs the algorithm 70 times, and computes the average result. To save time, we run each algorithm 10 times, and compute the average. 

The performance comparison of the four algorithms (random search, TPE, Hyperband, and Hyperband\_TPE) is presented in Figure \ref{fig:mnist}. The unit for the time axis is multiple of $R$ used (in this experiment, multiple of the 81 units), the same as in the Hyperband paper to make the comparison hardware independent. The reason that curves for Hyperband and our combination do not start at time 1 is that it takes some time for either algorithm to abandon all trial points but one, when the algorithm produces an output hyperparameter configuration.

Note that since this problem (i.e. train LeNet on MNIST) is an easy one, namely, there are many different hyperparameter configurations that can produce a trained LeNet with pretty high accuracy on MNIST, we see all four approaches quickly converges at a very high level (some even overlap). What we learn is that on simple problems, where good hyperparameter configurations reside massively in the searching space, no sophisticated hyperparameter optimization approach is needed. However, as we will show later in other experiments, when the problems get more complicated, more sophisticated approaches clearly have advantages over less sophisticated ones.

\subsubsection{AlexNet on Cifar10, MRBI, and SVHN}
The three experiments in this section is to train AlexNet respectively on three datasets: Cifar10 \cite{cifar10}, rotated MNIST with background images (MRBI) \cite{Larochelle:2007:EED:1273496.1273556}, and Street View House Number (SVHN) \cite{svhn}. There are eight hyperparameters: Initial learning rate, $l_{2}$ penalty for respectively three convolutional layers and one fully connected layer, learning rate reduction, and scale and power for local response normalization. We set the range of feasible values of each hyperparameter the same as that in the Hyperband paper. Clearly, these are harder problems than the previous one, in terms of complexity of neural network, hyperparameter searching space, and the datasets.

For Cifar10 and MRBI, $R$ = 300; for SVHN, $R$ = 600. $\eta=4$ for all three experiments. These numbers are the same as the Hyperband paper. Still, the number of training images corresponding to each unit of $R$ is set up empirically. These settings result into an equivalent number of about 18 trial points for either TPE or random search, where each trial point consumes all units (300 or 600) of $R$. We run each algorithm 10 times (the same as Hyperband paper this time), and compute the average accuracy.

\begin{figure}
\begin{minipage}[b]{1.0\linewidth}
  \centering
  \centerline{\includegraphics[width=0.6\textwidth]{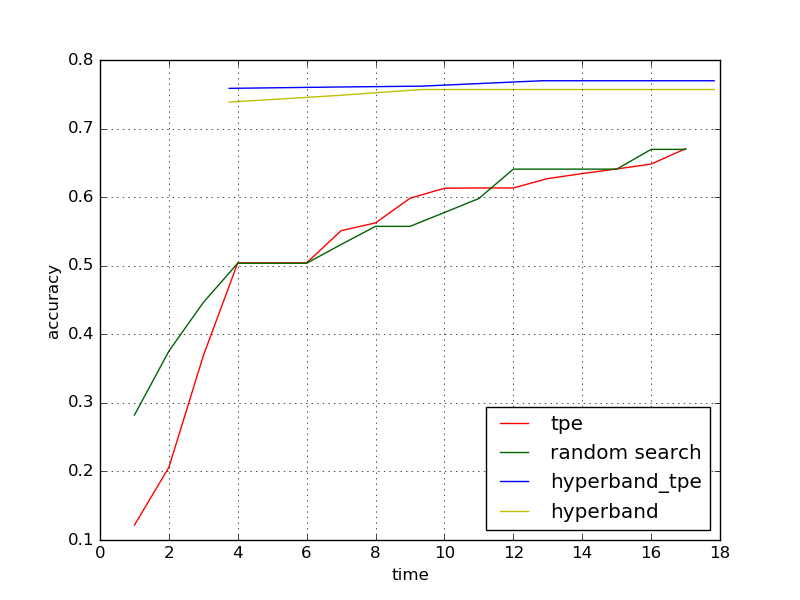}}
  \centerline{(a) Cifar10}\medskip
\end{minipage}

\begin{minipage}[b]{1.0\linewidth}
  \centering
  \centerline{\includegraphics[width=0.6\textwidth]{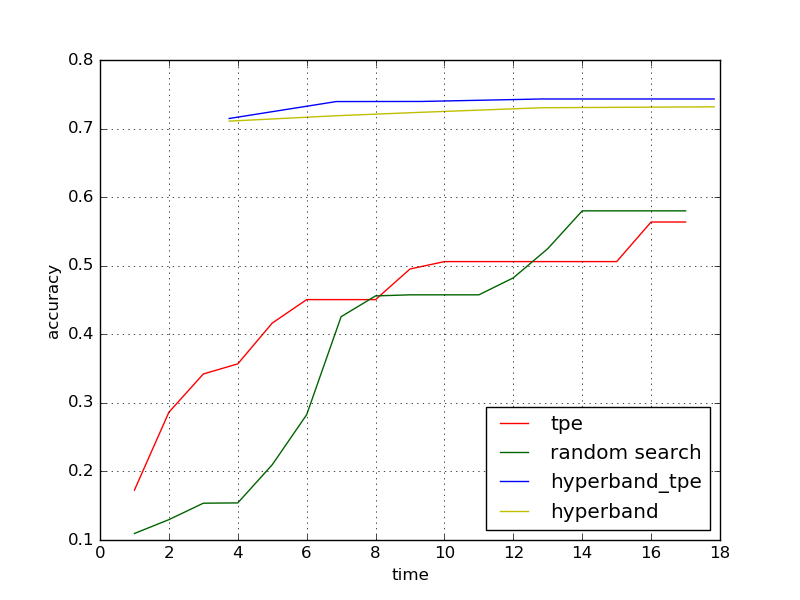}}
  \centerline{(b) MRBI}\medskip
\end{minipage}

\begin{minipage}[b]{1.0\linewidth}
  \centering
  \centerline{\includegraphics[width=0.6\textwidth]{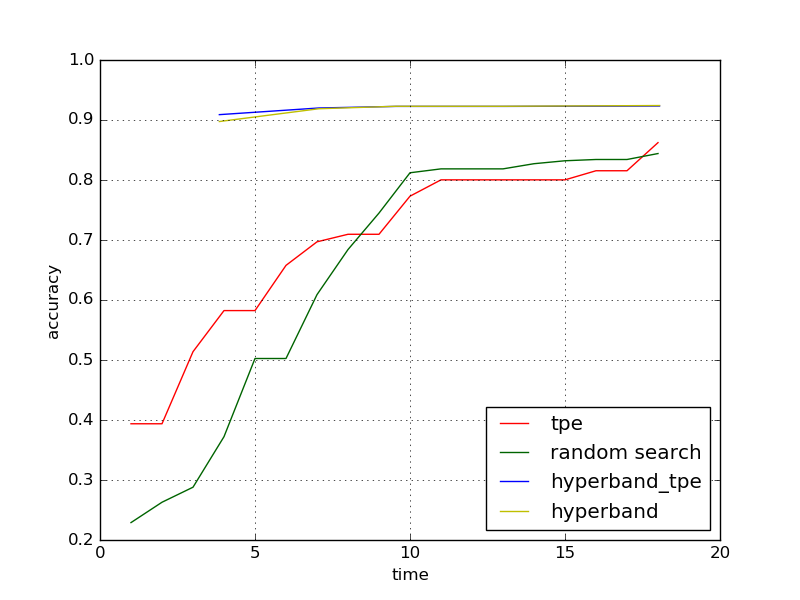}}
  \centerline{(c) SVHN}\medskip
\end{minipage}
  \caption{Average accuracy across 10 trials for AlexNet on Cifar10, MRBI, and SVHN.}
  \label{fig:alexnet}
\end{figure}

The performance comparison of the four algorithms (random search, TPE, Hyperband, and Hyperband\_TPE) is presented in Figure \ref{fig:alexnet}. Still, the unit for the time axis is multiple of $R$ used.

Unlike LeNet on MNIST, in these three problems, there are not as many hyperparameter configurations that can produce high accuracy, which is reflected by the separation between the curves. We see the two curves respectively for TPE and random search intertwine with each other, and the curve for Hyperband is much higher than the previous two. Although Hyperband forms a very strong baseline, yet our combination algorithm outperforms it slightly. We can also see that as the problem becomes harder and harder, from SVHN to Cifar10 then to MRBI (see the scale of the accuracy for these three problems), the gap between our combination algorithm and Hyperband also become larger. Note that although it might be possible that given sufficient or even infinite resources, all hyperparameter optimization algorithms may finally converge at a similar high accuracy, the target of the game here is to find a better solution as quickly as possible.

These results validate our idea of combining the strength of Hyperband and Bayesian optimization, and they also show that the combination algorithm has bigger advantage over other approaches, as the problems become more difficult.

\subsection{Decomposition of SSD}

In this section, our target problem is to find a low rank decomposed approximation neural network of the Single Shot Detector (SSD) \cite{DBLP:journals/corr/LiuAESR15}, to balance well between accuracy and speed. SSD is a fully convolutional neural network with 21 convolutional layers for objects detection. It achieves real time detection speed on high end GPU like Titan X, and competitive accuracy. However, on low end GPU like K520 on Amazon EC2 instances, real time detection cannot be achieved by SSD. Therefore, it is needed to speed up the computation, but probably at the cost of sacrificing certain accuracy.

The low rank regularization technique proposed in \cite{DBLP:journals/corr/TaiXWE15} decomposes both the shape and weights of each existing trained convolutional layer into some low rank approximation, by removing the redundant information contained in the tensor formed by all filters of each convolutional layer, via SVD. The produced low rank approximation is usually quicker than the original convolutional layer. Specifically, suppose for one convolutional layer, there are $N$ filters, each has $C$ channels, and is of size $d \times d$ spatially, then it is decomposed into two convolutional layers: First one has $K$ filters, $C$ channels, size $d \times 1$ spatially; and second one has $N$ filters, $K$ channels, size $1 \times d$ spatially. The hyperparameter $K$ controls the degree of information compression: The smaller the value of $K$ is, the more aggressively the layer is compressed, thus the quicker but less accurate the resulted approximation is.

We apply this low rank regularization technique on SSD (more specifically SSD300). Since SSD has 21 convolutional layers, our hyperparameter configuration space is of dimension 21, i.e., each dimension corresponds to the value of $K$ for one of the 21 convolutional layers, whose range is between 1 and the original number of filters in that layer. In this experiment, our objective function is not a single metric as in previous experiments, but a combination of two metrics, accuracy (in terms of map) and speed (in terms of fps): $\alpha \times map + fps$, where $\alpha$ is a parameter balancing the two metrics.

We test the accuracy of the network corresponding to each hyperparameter configuration on PASCAL VOC dataset \cite{Everingham10} (the tessellation of training and validation is the same as that in the SSD paper). Instead of directly reporting the accuracy of the decomposed approximation network, we actually finetune it for a small number of epochs (specifically 5) to allow it have a chance to adapt better, and report the accuracy after finetuning.

For Hyperband and our combination algorithm, we set $R$ = 2500, $\eta$ = 5, and the number of images corresponding to each unit of $R$ accordingly, to match the 5 epochs of finetuning. These settings result into an equivalent number of about 18 trial points for either TPE or random search, where each trial point consumes all 2500 units of $R$. We run each algorithm only once in this experiment in consideration of total running time.

\begin{figure}
  \centering
    \includegraphics[width=1.0\textwidth]{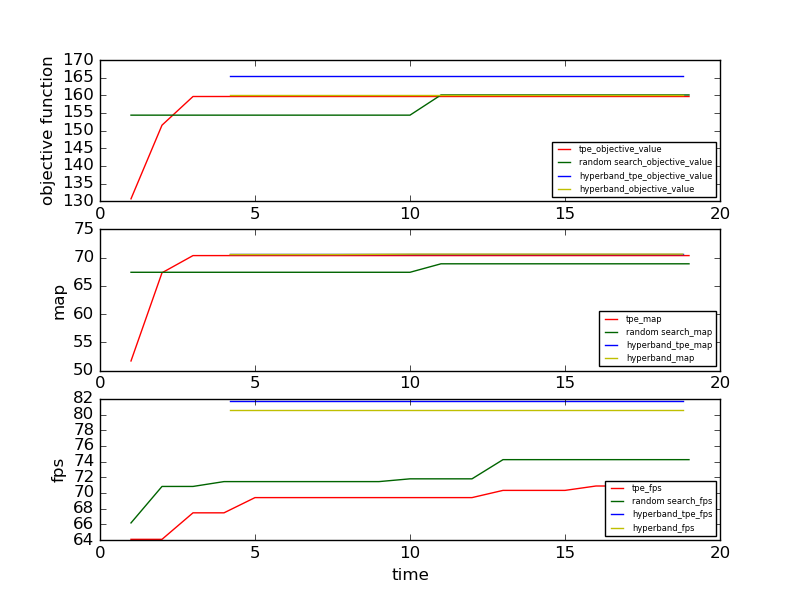}
  \caption{Objective function value, map, and fps for low rank decomposition of SSD.}
  \label{fig:ssd-low-rank}
\end{figure}

The performance comparison of the four algorithms (random search, TPE, Hyperband, and Hyperband\_TPE) is presented in Figure \ref{fig:ssd-low-rank}. Still, the unit for the time axis is multiple of $R$ used.

This time, we are facing a deeper network, higher dimensionality for hyperparameter configuration space, larger and more complex dataset, so it is not hard to imagine that configurations with good performance become even rarer (comparing to the size of the whole space). But again, our combination algorithm shows better performance than other approaches.

\section{Conclusion}

We proposed a novel approach for hyperparameter optimization in deep learning. Our approach combines the strength of both Hyperband algorithm and Bayesian optimization. We sampled trial points balancing between exploitation and exploration, while allocating more computational budget to more promising trial points. Experimental results validated that our combination approach could find better hyperparameter configuration more quickly than other approaches, such as Random search, Bayesian optimization, or Hyperband. Further more, the results also indicated that our approach outperforms other approaches by a larger margin, as the hyperparameter optimization problem become more complex and difficult.

%{\small
%\bibliographystyle{plain}
%\bibliography{mybib}
%}

\end{document}